# Deploy Large-Scale Deep Neural Networks in Resource Constrained IoT Devices with Local Quantization Region

Yi Yang, Andy Chen, Xiaoming Chen, Jiang Ji, Zhenyang Chen, Yan Dai
Intel Corporation

*Abstract*— Implementing large-scale deep neural networks with high computational complexity on low-cost IoT devices may inevitably be constrained by limited computation resource, making the devices hard to respond in real-time. This disjunction makes the state-of-art deep learning algorithms, i.e. CNN (Convolutional Neural Networks), incompatible with IoT world. We present a low-bit (range from 8-bit to 1-bit) scheme with our local quantization region algorithm. We use models in Caffe model zoo as our example tasks to evaluate the effect of our low precision data representation scheme. With the available of local quantization region, we find implementations on top of those schemes could greatly retain the model's accuracy, besides the reduction of computational complexity. For example, our 8-bit scheme has no drops on top-1 and top-5 accuracy with 2x speed-up on Intel Edison IoT platform. Implementations based on our 4-bit, 2-bit or 1-bit scheme are also applicable to IoT devices with advances of low computational complexity. For example, the drop on our task is only 0.7% when using 2-bit scheme, a scheme which could largely save transistors. Making low-bit scheme usable here opens a new door for further optimization on commodity IoT controller, i.e. extra speed-up could be achieved by replacing multiply-accumulate operations with the proposed table look-up operations. The whole study offers a new approach to relief the challenge of bring advanced deep learning algorithm to resource constrained low-cost IoT device.

## I. Introduction

Deep neural networks seem to be quite promising in recent years and have been demonstrated their power in many applications, like visual object recognition and classification (Ciresan et al., 2011; Krizhevsky et al., 2012; Sermanet et al., 2013), speech recognition (Hinton et al., 2012; Mohamed et al. 2012) and language processing (Collobert et al., 2011). The advances of deep learning algorithm are making machines to understand the world and facilitate people better. There is evidence that scaling up deep learning algorithm, including larger amount of training data, more model parameters could increasingly improve its performance (Coates et al., 2010; Coates et al. 2013) due to more supervision information absorbed. Therefore, training extremely large network model using high performance computing infrastructure may evidently lead to higher intelligence. Some notable work like using thousands of CPU cores to train a deep network with billions of parameters (Dean et al. 2012), using clustered machines to train a model with 1 billion connections (Le et at., 2013), using high-end GPUs (Coates et al. 2013). Therefore, such large scale learning algorithm could be extremely computationally expensive.

In the meanwhile, mobile devices and IoT devices are tend to be more intelligent equipped by increasing amount of machine learning applications. This is leading to a great surge of interest in deployment of deep neural networks in embedded systems. However, the expensive computational cost and heavy memory overhead make it unaffordable, addressing much work to this dilemma. One common solution is to transmit the runtime data to a server center and send back the processed data back to IoT devices (Chun et.al. 2009). However, this could lead to further issues, like network traffic issues or privacy issues (Yuan et.al. 2013). Memory overhead introduced by large scale of deep neural networks is explored by (Denil et al., 2013), in which large amounts of redundancy in neural networks are demonstrated, leading further work on neural networks compression (Ba et al., 2013; Gong et al., 2014). However, the decompression of compressed neural networks may impose extra burden on computation ability in another way. Solving bottleneck on computational complexity is another focus. Researchers propose to use additional hardware accelerators implemented on FPGA or ASIC devices (Peter et al., 2008; Kim et al., 2009; Gokhale et al., 2014), which are believed to be more power efficient than general purpose hardware due to its specifically customized hardware based on deep learning algorithm. The proposed hardware accelerator is based on FPGA whose performance per watt is six times that of most mobile processors and twenty times that of high-end CPUs or GPUs (Gokhale et al., 2014). Though power effective, hardware accelerators might not be cost effective due to extra chip area, i.e. scale of programmable logic array increases with neural networks. In summary, deployment of deep neural networks in resource constrained IoT devices may still face serious computation bottleneck or area cost.

In this article, we focus on fixed point implementation and propose one novel quantization approach called "local based quantization". Our target is to deploy the mainstream deep neural networks on resource constrained IoT devices by limiting the numeric representation to an extremely low precision while still ensure high task quality. We are to study how to eliminate quantization errors in a further method if the neurons in the deep neural networks are extremely low represented.

The objective of this article is to explore the possibility of deploying deep neural networks in resource constrained IoT devices with extremely limited area budget and computation resources. This article will be organized as follows:

- In section II, related work will be given and originality of our work will be demonstrated briefly.
- In section III, the process of typical forward propagation in deep learning algorithm will be demonstrated, together with the limitation of its conventional 32-bit floating point implementation.
- In section IV, the proposed local based quantization scheme in deep neural networks is to be illustrated in detail, together with the comparison with the prior quantization scheme.
- In section V, one look-up table scheme is to be proposed to further reduce computational complexity.
- In section VI, extensive experiments are to be demonstrated for evaluation, including speedup, task quality and hardware resource on FPGA.
- In section VII, we will summarize our work as the final conclusion.

## II. RELATED WORK

There are several related work focusing on quantization of deep neural networks with low precision (Vanhoucke et al., 2011; Courbariaux et al., 2014). Vanhoucke et al. use 8-bit linear quantization to convert activations into unsigned char and weights into signed char, by which the performance of a speech recognition task is evaluated on x86 platform (Vanhoucke et al., 2011). Courbariaux et al. further explored the possibility to train models with low precision, where 10 bits are used for storing activations and 12 bits for storing parameters of Maxout networks (Courbariaux et al., 2014). Our work is nevertheless focusing on deploying neural network in resource constrained IoT devices, with the help of extremely low precision, where quantization scheme of 4bits and 2 bits are investigated. The proposed new approach can avoid the errors introduced by extremely low precision. Our experiments are not only limited to task accuracy but also focused on speedup and hardware resources.

## III. TYPICAL DEEP NERUAL NETWORKS IMPLEMENTATION

In this section we present a brief knowledge of typical deep neural network structure and its forward propagation, which is the data path of tasks like image classification or speech recognition. Together, inferiorities of the conventional floating point implementation are to be described.

### A. Typical deep nerual networks structure

Figure.1 shows one typical fully-connected deep neural networks, consisting of multiple layers and all neurons in each layer are connected to each neuron in next layer.

The forward propagation could be defined as follows: visible inputs are propagated forward through each layer and after transformation with transition function using activation functions like sigmoid or ReLU (Nair et al., 2010), then outputs of this layer are propagated forward as the inputs of the next layer. The overall forward procedure could be described as:

$$for\ l\ in\ layers:\ \alpha_i^{l+1} = f\left(\sum_{j=0} W_{ij}^l \alpha_j^l\right) \quad (1)$$

Where $W^l$ is the weight matrix of layer $l$ and $W_{ij}^l$ is the $(i,j)^{th}$ value of matrix $W^l$. The function above denotes the activation of each layer. $\alpha^l$ is the vector which denotes the neurons of each layer, including hidden layer and visible layer and $\alpha_j^l$ is its $j^{th}$ element.

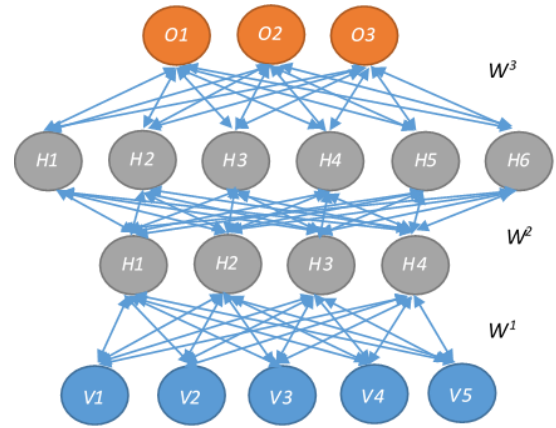

Figure.1 Illustration of typical fully-connected deep neural networks

### B. Original floating point implementation

For now, there are three floating point formats supported, which are half precision, single precision and double precision floating point, as illustrated in Table 1.

Table 1 Supported floating point formats definition

|  | Total | Exponent (E) | Mantissa (M) |
|---|---|---|---|
| Double precision | 64 bits | 11 bits | 52 bits |
| Single precision | 32 bits | 8 bits | 23 bits |
| Half precision | 16 bits | 5 bits | 10 bits |

Original implementations of deep learning algorithms typically use 32-bit single point floating-point type to represent the real values of all data in various layers. The exponent and mantissa parts determine the representation range and precision respectively, by which the floating point value is:

$$value_{sp} = (-1)^{sign} \times \left(1 + \frac{mantissa}{2^{23}}\right) \times 2^{(exponent-127)} \quad (2)$$

### C. Limiation of floating point implementation on IoT devices

Using floating point based implementation could improve task accuracy to some extent, but the limitation is also obvious. Memory storage and related access bandwidth could be a heavy burden for high precision representation. For example, the deep CNN (convolution neural network) proposed by Alex Krizhevsky (Krizhevsky et.al., 2012) contains eight weights layers (the first five are convolutional and the remaining three fully-connected layers). There could be hundreds of million neurons altogether.

Computational parallelism is another limitation. Equation (1) shows the key operation in each layer is multiply-accumulate on vectors. Typically, modern embedded processor or low power graphic IPs leverage fixed width SIMD lanes to process data in parallel. For example, ARMv8 provides 128-bit vector register and use '*fmla*' instruction for multiply-accumulate operation which can support four 32-bit single precision floating point figures execution as

*fmla v1.4s, v21.4s, v2.4s*

Memory bandwidth and SIMD throughput are constrained resource for IoT device. Computation throughput decreases linearly with bit-width of one figure.

## IV. FIXED POINT IMPLEMENTATION WITH LOCAL BASED QUANTIZATION

In this section we present one fixed point implementation using our local based quantization scheme. The proposed scheme could ensure high accuracy even when precision of numeric representation is extremely low.

### A. Quantization effect in fixed-point implementation

Converting a floating-point numeric representation into a fixed-point one may inevitably introduce quantization errors. This could be a disaster especially using low precision, and result in a fatal error by an activation function, like ReLU or sigmoid function. One quantized value of '*x*' with round-to-nearest quantization scheme $Q(x)$ could be expressed as:

$$Q(x) = round\left(\frac{x - x_{min}}{s}\right) \quad (3)$$

The '*x*' and '*s*' in equation (3) are original floating-point value and quantization step. Therefore, the quantization error could be expressed as:

$$e_{Q(x)} = x - Q(x) \quad (4)$$

If to convert floating-point values into fixed-point ones using '*n*' bit quantization and normalize a range of [$x_{min}$, $x_{max}$] floating-point values into a full quantization range at the same time, the fixed-point arithmetic quantization curves and its error curves are shown in figure.2:

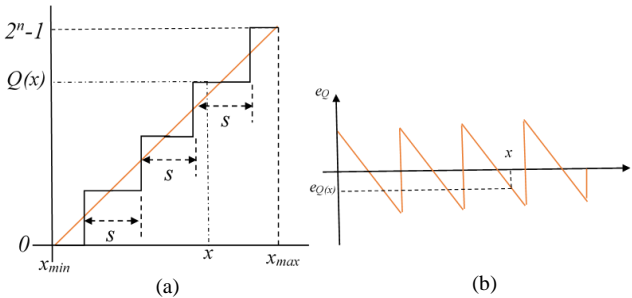

Figure.2 (a) Fixed-point arithmetic quantization curves and the (b) error curves

It is clear from the curves that the errors introduced by quantization are determined by quantization step, which is expressed by:

$$s = \frac{x_{max} - x_{min}}{2^n - 1} \quad (5)$$

Therefore, making the quantization step as small as possible by grouping those weight neurons with similar numeric values within the same quantization region is beneficial to eliminate quantization errors.

### B. Dynamic fixed point in neural networks

Prior quantization scheme proposed for fixed point deep learning algorithm, like dynamic fixed point (Courbariaux et al., 2014), uses dynamic scaling factors or quantization steps, which are various from layer to layer to better meet the different numeric range. The forward propagation expressed in equation (1) turns to:

$$\alpha_i^{l+1} = f\left(\sum_{j=0} W_{ij}^l \otimes \alpha_j^l\right)$$

$$= f\left(Q_l^{-1} \sum_{j=0} Q_l(W_{ij}^l) \otimes \alpha_j^l\right)$$

$$s_l = \frac{x_{max}^l - x_{min}^l}{2^n - 1} \quad (6)$$

$$Q_l(x) = round\left(\frac{x - x_{min}}{s_l}\right)$$

The '$Q_l(x)$' is the quantization function in layer '*l*' which converts one floating point value '*x*' into fixed point one. And '$Q_l^{-1}(x)$' is the dequantization function in layer '*l*' which converts fixed point value back to floating point one. Each layer's quantization function shares one scaling factor, various from layer to layer. The structure is illustrated in Figure.3.

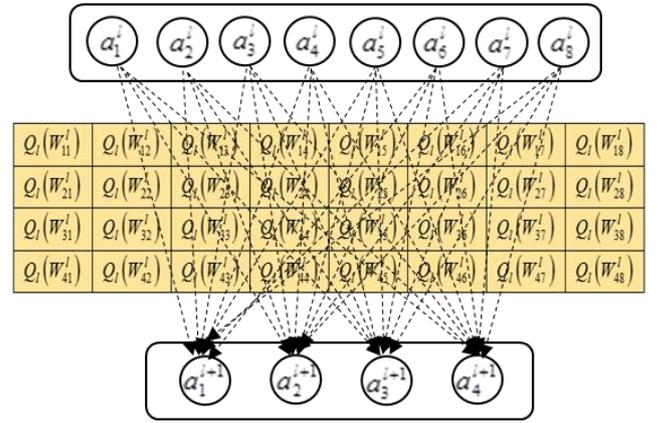

Figure.3 Illustration of dynamic fixed point in neural networks

### C. Local based quantization scheme in neural networks

The dynamic fixed point scheme works well when quantization precision is about 8 ~ 12 bits, with no accuracy drops. However, further low numeric precision, like 6 bits, 4 bits or even 2 bits, may be infeasible due to the large quantization error accumulated from layer to layer.

To alleviate such issues, we proposed one local based quantization scheme. The key idea is to use smaller quantization regions, whose scaling factor or quantization step is locally shared, instead of globally shared by all neurons within one layer as prior schemes.

The proposed local based quantization is illustrated in Figure.4. Different from Figure.3, the weights layer '*l*' is divided into 4 local quantization regions. Each local quantization region has 8 elements sharing one same quantization step, determined by the maximum value $x^k_{max}$ and minimum value $x^k_{min}$ within the region *k*. This local quantization step is typically smaller than the global quantization step. The forward propagation could be expressed as:

$$\alpha_i^{l+1} = f\left(\sum_{j=0} W_{ij}^l \otimes \alpha_j^l\right)$$
$$= f\left(\sum_{k=0}^{n} \left(Q_{lk}^{-1} \sum_{j=0}^{N/n} Q_{lk}(W_{ij}^l) \otimes \alpha_j^l\right)\right)$$

$$s_{lk} = \frac{x_{max}^{lk} - x_{min}^{lk}}{2^n - 1}$$

$$Q_{lk}(x) = round\left(\frac{x - x_{min}}{s_{lk}}\right)$$
(7)

The variable '*N*' in equation denotes the number of elements in the quantized region and '*n*' is the number of local quantization regions. Therefore each local region has '*N/n*' quantized elements.

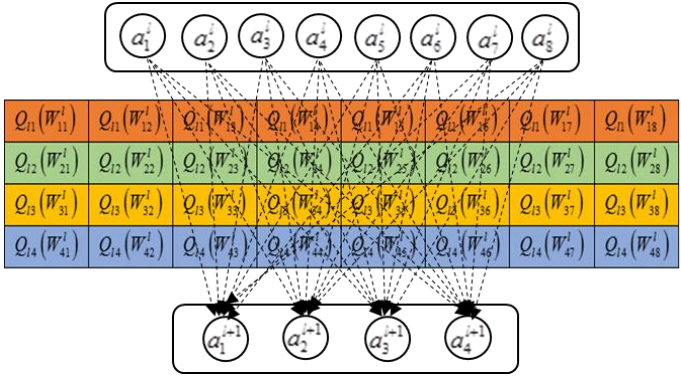

Figure.4 Local based quantization in neural networks (four colors denote four local regions)

Observed from equation.6 and equation.7, more operations are needed in local based quantization since every maximum value and minimum value must be determined in the local quantization region. However, the proposed scheme could reduce overall computation complexity per operation largely due to the higher level's data parallelism coming from with the advances of lower-bit numeric representation.

## V. COMBINATION OF LOOK-UP TABLE SCHEME WITH LOCAL BASED QUANTIZATION

In this section, we present one look-up table scheme combined with previous local based quantization, which is to further optimize the performance when deploy large-scale deep neural networks in resource constrained IoT devices.

### A. Overview

Quantization errors could be reduced if the size of local region becomes smaller when using local based quantization. With a proper region size, extremely low precision could provide high performance without accuracy drops in one image recognition task, like 8-bit, 6-bit, 4-bit or even 2-bit implementation.

However, since there is no standard ISA support below 8-bit format, data parallelism in general CPUs could not be achieved using lower precision even if they are perfect enough for implementation. Therefore, further optimization could be made using the proposed look-up table. By store the indexed values in one look-up table, multiply-accumulate operations can be saved, thus boosting the performance of deep learning algorithms on IoT devices. Due to the previous local based quantization scheme, extremely low precision can be achieved, thus keeping the size of look-up table relative small.

### B. Look-up table implementation

The overall structure of look-up table implementation is given in Figure.5. The previous multiply-accumulate operations are replaced by add operations by storing indexed values in the table. The table size is determined by the quantization precision and kernel size.

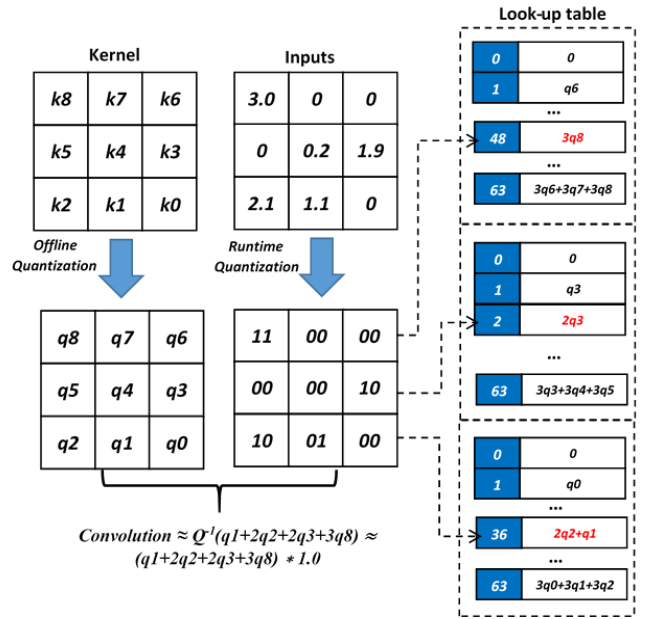

Figure.5 combination of 2-bit quantization with look-up table scheme

The inputs and kernel are both quantified from floating point representation into fixed point one. However, the quantization of kernels in trained deep neural networks is done offline while the inputs have to be converted into fixed point in runtime.

### C. Forward propagation with look-up table scheme

There is no multiply operations when using a look-up table scheme, and the table size is relatively small if the quantization precision is low enough as illustrated in Figure.5. Therefore, large portion of tedious multiply-accumulate operations are saved, thus boosting the performance in a certain extent. The forward propagation is expressed as:

$$\alpha_i^{l+1} = f\left(\sum_{j=0} W_{ij}^l \alpha_j^l\right)$$
$$= f\left(\sum_{k=0}^n Q_{lk}^{-1}\left(Index\left[\alpha^l\right]\right)\right) \quad (8)$$

Vector '$\alpha^l$' denotes the input vector of layer '$l$' and is used to index the look-up table, where various quantified sums of neurons are stored, as shown in Figure.5.

## VI. EVALUATION

In this section, we conduct thorough experiments to evaluate our local based quantization scheme and the derivative look-up table scheme, from both performance improvements and cost saving sides. The experiments include the comparison between 32-bit floating point, 8-bit fixed point, 6-bit fixed point, 4-bit fixed point and 2-bit fixed-point, mainly focused on example task performance, computational complexity and device cost.

### A. The example task and dataset

Our example task is to deploy large-scale deep convolutional neural networks from Caffe model zoo on the resource constrained IoT device, to classify the 1.2 million high-resolution images into 1000 various classes, using 50,000 images as validation and 150,000 images as testing images. The dataset is from ImageNet LSVRC-2012 contest (Deng et.al. 2012), which is a subset of ImageNet. ImageNet is a dataset of over 15 million labeled high-resolution images, classified into 22,000 categories.

The trained deep neural networks we are using are two popular deep algorithms: 'AlexNet' (Krizhevsky et.al. 2012) from ImageNet LSVRC-2012 contest and 'VGG' (Simonyan et.al. 2014) from ImageNet LSVRC-2014 contest. 'AlexNet' is an 8 weight layer neural network while 'VGG' has various variants. We chose the one with 16 weight layers and all receptive field is 3x3.

### B. Hardware platform

Our experiments were all implemented on Intel Edison IoT board, which is powered by Silvermount architecture and 1GB RAM. We implemented 32-bit floating point scheme as the baseline, which offloads matrix correlation based convolution operation to Intel Math Kernel Library (MKL) for best hardware efficiency.

### C. Software platform

The software platform we were using for study is called '**BLAImark**', which is a framework we used to evaluate neuron network oriented algorithms on various platforms. The framework is illustrated in Figure.6. The deep neural networks are supplied and quantified offline using the proposed local quantization scheme. The image datasets are inputted and quantified during runtime, operated with the pre-quantified networks from layer to layer until the final results are outputted as accuracy and performance reports.

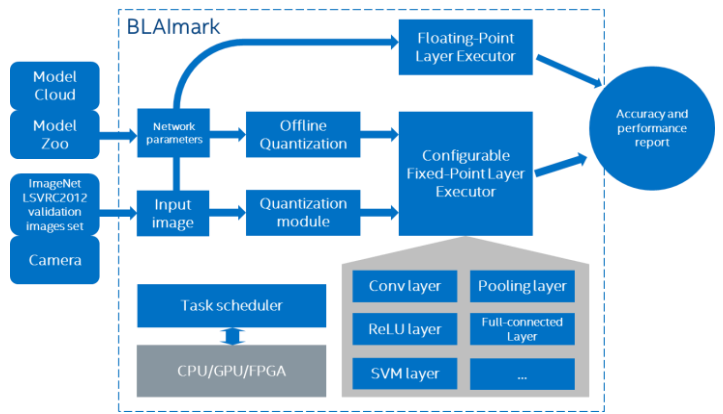

Figure.6 BLAImark software platform for deep learning study

### D. Performance gain from fixed point implementation

We implemented the 'AlexNet' and 'VGG' based fixed point CNN algorithm with 8-bit quantization precision and evaluate the speedup and accuracy drop compared with the original MKL baseline. To ensure high accuracy, we use the proposed local based quantization scheme here and select the size of quantization region as large as the kernel size, which is typically quite large. For example, the first convolution layer in 'AlexNet' filters the 224×224×3 input image with 96 kernels of size 11×11×3 with a stride of 4 pixels, as illustrated in Figure.7. Therefore, for this first convolution layer, we quantified the kernel offline and inputs at runtime both into 8-bit fixed point numeric representation, with a local quantization region of 363 (11×11×3). And the involved GEMM (general matrix multiply) is $M*M$, where $M \in R^{11\times 11}$.

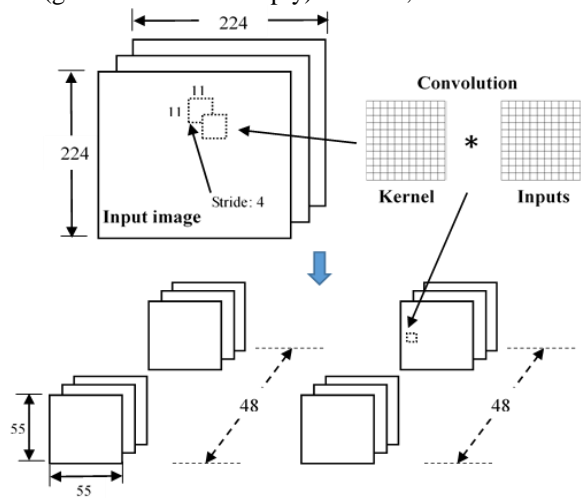

Figure.7 Illustration of first convolution layer in 'AlexNet'

The speedup chart is given in Figure.8. Top-1 and top-5 accuracies of the 1000 categories classification task are listed in Table.1. The vertical axis in Figure.7 denotes the runtime to recognize one image using the deep neural network of horizon axis**.**

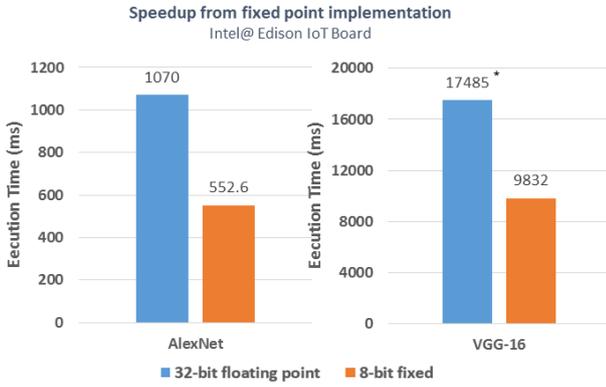

Figure.8 Chart of speedup from fixed point

Table.1 Accuracy comparison of top-1 and top-5 test between baseline and 8-bit quantization

|  | AlexNet | | VGG-16 | |
| --- | --- | --- | --- | --- |
|  | top-1 | top-5 | top-1 | top-5 |
| **32-bit floating** | 56.6% | 80.0% | 68.9% | 88.3% |
| **8-bit fixed** | 56.6% | 80.0% | 68.6% | 88.2% |

From Figure.8 and Table.1, we can see our version fixed point scheme could boost the overall image recognition task by about 2 times, with quite limited accuracy drop for both top-1 and top-5 test results and for both deep neural networks.

*E. Accuray improvemnts from local based quantization*

Here we conducted experiments on quantization with further low precision, like 6-bit, 4-bit and 2-bit. We were to test how local based quantization could benefit the accuracy improvements with low-precision scheme by reducing quantization errors.

Test accuracy for 'AlexNet' and 'VGG-16' are listed in Table.2, including original 32-bit floating point baseline, 8/6/4/2-bit dynamic fixed point quantization (Section IV, part B) and the proposed 8/6/4/2-bit local based quantization scheme. All the weights parameters, like convolution kernels, are quantified offline into static 8-bit fixed point, while the inputs data are various from 8-bit to 2-bit precision. Note '**DQ**' is short for dynamic fixed point quantization and '**LQ**' is short for local based quantization.

Table.2 Accuracy drop with various quantization precision

|  |  | 8-bit | 6-bit | 4-bit | 2-bit |
| --- | --- | --- | --- | --- | --- |
| **AlexNet** | DQ top-1 | 56.5% | 56.4% | 55.2% | **22.9%** |
|  | LQ top-1 | 56.6% | 56.6% | 56.4% | **46.8%** |
|  | DQ top-5 | 79.9% | 79.8% | 78.7% | **42.8%** |
|  | LQ top-5 | 80.0% | 80.0% | 79.8% | **71.5%** |
| **VGG-16** | DQ top-1 | 68.7% | 68.7% | 63.7% | **1.5%** |
|  | LQ top-1 | 68.6% | 68.6% | 68.2% | **50.2%** |
|  | DQ top-5 | 88.2% | 88.2% | 85.0% | **4.4%** |
|  | LQ top-5 | 88.2% | 88.2% | 88.1% | **74.6%** |

The corresponding charts of 'AlexNet' and 'VGG-16' are listed in Figure.9, respectively. The '**DQ**' chart and '**LQ**' are represented by dash line and solid line respectively. From the chart, it is obviously observed that by using the proposed local based quantization scheme, the accuracy is extensively improved for both 'AlexNet' and 'VGG-16', also for all the tested quantization accuracy, especially for 2-bit quantization. The accuracy is improved from 22.9% to 46.8% for 'AlexNet' top-1 test using 2-bit quantization, and improved from 1.5% to 50.2% when deploying 'VGG-16'.

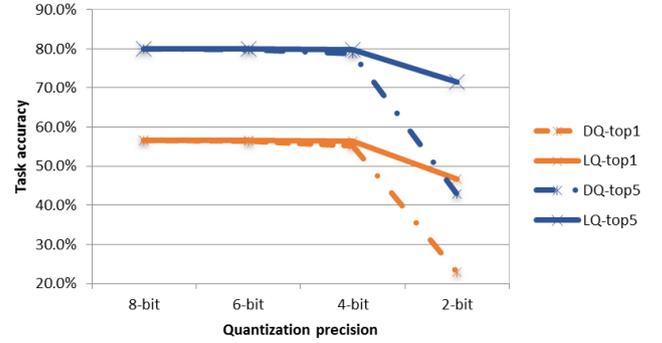

(a)

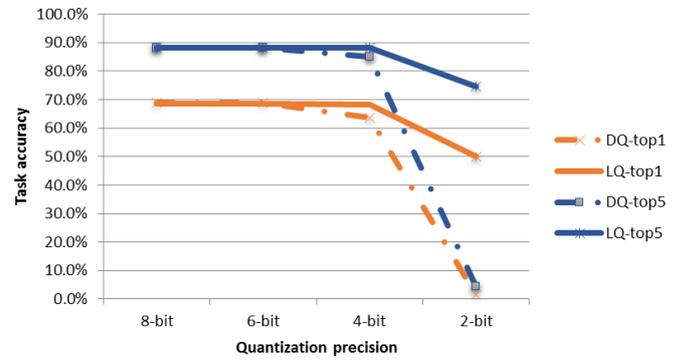

(b)

Figure.9 Task accuracy with various quantization precision on (a) 'AlexNet' and (b) 'VGG-16'

It proves the proposed the quantization scheme could be beneficial to realize the application of advanced deep learning algorithms to resource constrained IoT devices.

*F. Improve accuracy by smaller local quantization region*

All the previous experiments implemented the proposed local based quantization scheme by dynamically choosing the local region with the same size of the kernel size, which are typically very large in common deep neural networks. However, by empirically using a smaller local quantization region, further accuracy could be achieved when the precision is extremely low as 2-bit.

To evaluate this idea, we conducted further experiments using 2-bit on 'VGG-16' and the results are shown in Figure.10. Obviously, using smaller local quantization region could further improve task accuracy when numeric precision is extremely limited. Our experiments show that the task accuracy was improved from 50.2% to 68.3% for top-1 and 74.6% to 88.0% for top-5.

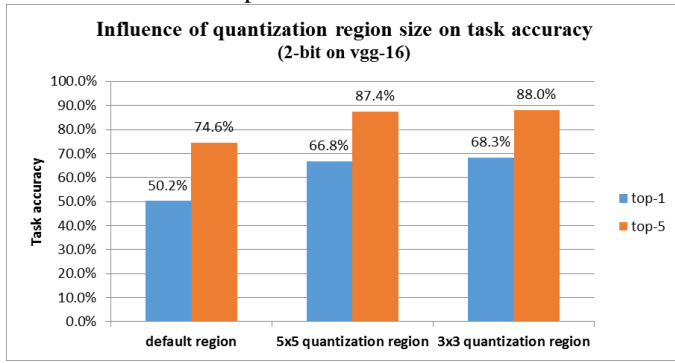

Figure.10 Influence of quantization region on task accuracy

*G. Reduce computational complexity using look-up table*

We conducted experiments to evaluate the contribution of look-up table to computational complexity reduction. We limited the inputs precision to 2 bits and fix the weights parameters to 8 bits. The statistical results of calculation amount of convolutional layers to process one input image are listed in Table.3.

Table.3 Multiply and add operations statistical results

| Network | Scheme | Multiply (M) | Add (M) |
|---|---|---|---|
| AlexNet | original | 666 | 666 |
|  | 2-bit LUT | 74 | 222 |
| VGG-16 | original | 15347 | 15347 |
|  | 2-bit LUT | 1705 | 5116 |

*H. FPGA based hardware resource evaluation*

To further evaluate the benefit of our extremely low quantization scheme on cost and power, we built one Matrix Multiplier using FPGA. Since the multiply-accumulate operations are the most frequent operations in one deep learning algorithm, a Matrix Multiplier is the key component of the hardware design and would cost most chip area. The top architecture is shown in Figure.11, where '**ISC**' denotes "Input Stream Controller", '**PSC**' denotes "Parameter Stream Controller", '**CU**' is the "Computing Unit", '$W_p$' is the width of weight parameters and '$W_i$' is the width of input elements.

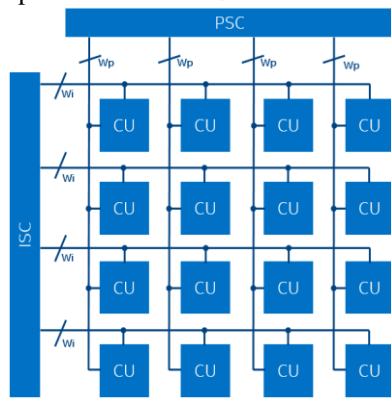

Figure.11 top architecture of Matrix Multiplier

The **CU** in Figure.11 is actually a multiply-accumulator which will multiply and accumulate a series of data and generate proper output, as shown in Figure.12. Our Matrix Multiplier has 4x4 **CU**. To complete the multiplication operation of input matrix and parameter matrix, each **CU** receives two elements from input matrix and parameter matrix, scheduled by **ISC** and **PSC**.

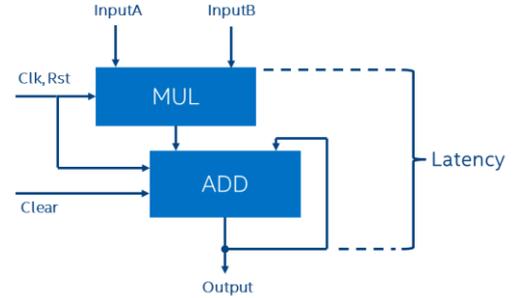

Figure.12 CU design

We built our Matrix Multiplier on "Xilinx XC6vlx240t-1ff1156" using "ISE 13.4" for synthesis. Below are several specs of the Matrix Multiplier, including hardware resources in Table.3 and performance evaluation in Table.4. Note the '**Fixed 8×n**' in Table.4 and Table.5 means that the weight width '$W_p$' is 8-bit fixed point and inputs width '$W_i$' is quantified to *n* bits.

Table.4 Hardware resources of Matrix Multiplier module

| Configuration | LUT # | FF # | Max Freq | Latency |
|---|---|---|---|---|
| FP 32×32 | 17534 | 11586 | 269 MHz | 8 |
| Fixed 8×8 | 1571 | 1442 | 322 MHz | 3 |
| Fixed 8×4 | 923 | 962 | 532 MHz | 3 |
| Fixed 8×2 | 535 | 562 | 556 MHz | 2 |

Table.5 Performance and power evaluation

| Configuration | Performance @ Max Freq @ 90% utilization of LUTs [1] | Power @ 200MHz (clock/logic/signal power) [2] |
|---|---|---|
| FP 32×32 | 67 Gflops | 643 mW |
| Fixed 8×8 | 890 Gops | 71 mW |
| Fixed 8×4 | 2502 Gops | 51 mW |
| Fixed 8×2 | 4511 Gops | 37 mW |

*Note 1: performance is measured when all the configurations utilize same FPGA resources (90% utilization of all)
*Note 2: power is measured for a single matrix multiplier at 200MHz, whose hardware resource is according to the specific configuration

Obviously, the proposed quantization scheme helps to reduce the hardware cost when deploying large-scale deep neural networks to low cost IoT devices or specific hardware accelerator by retaining high accuracy with extremely limited numeric precision. This is believed to be significantly beneficial to accelerate the applications of advanced deep learning algorithms to IoT world.

## VII. CONCLUSION

The work of this article is to accelerate the applications of advanced deep learning algorithms to controller and gateway.

The key challenge is to deploy the mainstream deep neural networks, which are typically large-scale, to resource constrained devices with high performance, including execution time, task accuracy and costs.

We evaluated current approaches and proposed a new solution to enable extremely low precision implementation, including local based quantization, look-up table scheme and further errors elimination method by smaller quantization region.

We conducted extensive experiments to evaluate our scheme by deploying two popular deep neural networks, 'AlexNet' and 'VGG-16' to IoT devices. We used original 32-bit floating point implementation as baseline and demonstrated how our scheme help to boost the speed of example task by reducing computational complexity i.e. the overall task speed is boosted by a factor of 2 on Edison IoT board. We illustrated how our scheme help to retain high task accuracy even when the precision is limited to 2-bit, i.e. the top-1 accuracy is improved from 1.5% to 50.2% for 'VGG-16' and further improved to 68.3% using smaller quantization region. Our FPGA based experiments demonstrated how our scheme contributes to hardware saving and improving overall performance with lower power consumed.

REFERENCES

[1] Ciresan D C, Meier U, Gambardella L M, et al. Deep Big Simple Neural Nets Excel on Handwritten Digit Recognition[J]. Corr, 2010.
[2] Krizhevsky, Alex, Sutskever, Ilya, and Hinton, Geoffrey E. ImageNet Classification with Deep Convolutional Neural Networks [J]. Advances in Neural Information Processing Systems, 2012:2012.
[3] Sermanet P, Eigen D, Zhang X, et al. OverFeat: Integrated Recognition, Localization and Detection using Convolutional Networks [J]. Eprint Arxiv, 2013.
[4] Hinton G, Deng L, Yu D, et al. Deep Neural Networks for Acoustic Modeling in Speech Recognition[J]. IEEE Signal Processing Magazine, 2012, 29(6):82 - 97.
[5] Mohamed A, Dahl G E, Hinton G. Acoustic Modeling Using Deep Belief Networks [J]. IEEE Transactions on Audio Speech & Language Processing, 2012, 20(1):14--22.
[6] Collobert R, Weston J, Bottou L, et al. Natural Language Processing (almost) from Scratch [J]. Journal of Machine Learning Research, 2011, 12(1):2493-2537.
[7] Coates A, Lee H, Ng A Y. An Analysis of Single-Layer Networks in Unsupervised Feature Learning. [J]. In Aistats, 2011, 15.
[8] Coates A, Huval B, Wang T, et al. Deep learning with COTS HPC systems [J]. Proceedings of International Conference on Machine Learning, 2013:1337-1345.
[9] Dean J, Corrado G S, Monga R, et al. Large Scale Distributed Deep Networks [J]. Advances in Neural Information Processing Systems, 2012.
[10] Le Q V, Ranzato M, Monga R, et al. Building high-level features using large scale unsupervised learning [J]. International Conference on Machine Learning, 2011:8595 - 8598.
[11] Denil M, Shakibi B, Dinh L, et al. Predicting Parameters in Deep Learning [J]. Eprint Arxiv, 2013:2148-2156.
[12] Gong Y, Liu L, Yang M, et al. Compressing Deep Convolutional Networks using Vector Quantization [J]. Eprint Arxiv, 2014.
[13] Ba L J, Caruana R. Do Deep Nets Really Need to be Deep? [J]. Advances in Neural Information Processing Systems, 2013:2654-2662.
[14] Peter H, Srihari G, Durdanovic C I, et al. A Massively Parallel Digital Learning Processor [J]. Advances in Neural Information Processing Systems, 2008.
[15] Kim S K, Mcafee L C, Mcmahon P L, et al. A highly scalable Restricted Boltzmann Machine FPGA implementation [J]. International Conference on Field Programmable Logic & Applications, 2009:367 - 372.
[16] Gokhale V, Jin J, Dundar A, et al. A 240 G-ops/s Mobile Coprocessor for Deep Neural Networks[C]. //IEEE Conference on Computer Vision & Pattern Recognition Workshops. IEEE Computer Society, 2014:696-701.
[17] Vanhoucke V, Senior A, Mao M Z. Improving the speed of neural networks on CPUs [J]. Deep Learning & Unsupervised Feature Learning Workshop Nips, 2011.
[18] Courbariaux M, Bengio Y, David J P. Low precision storage for deep learning [J]. Eprint Arxiv, 2014.
[19] Nair V, Hinton G E. Rectified Linear Units Improve Restricted Boltzmann Machines.[J]. Proc Icml, 2010:807-814.
[20] J. Deng, A. Berg, S. Satheesh, H. Su, A. Khosla, and L. Fei-Fei. ILSVRC-2012, 2012. URL http://www.image-net.org/challenges/LSVRC/2012/.
[21] Simonyan K, Zisserman A, Simonyan K, et al. Very Deep Convolutional Networks for Large-Scale Image Recognition [J]. Eprint Arxiv, 2014.
[22] Chun B G, Maniatis P. Augmented Smartphone Applications Through Clone Cloud Execution [C] //HotOS. 2009, 9: 8-11.
[23] Yuan J, Yu S. Privacy Preserving Back-Propagation Neural Network Learning Made Practical with Cloud Computing [J]. Parallel & Distributed Systems IEEE Transactions on, 2013, 25(1):1.